\documentclass[conference]{IEEEtran}
\IEEEoverridecommandlockouts
\usepackage{cite}
\usepackage{amsmath,amssymb,amsfonts}
\usepackage{algorithmic}
\usepackage{graphicx}
\usepackage{textcomp}
\usepackage{xcolor}

\usepackage{subcaption}
\usepackage{url}            
\usepackage{booktabs}       
\usepackage[hidelinks]{hyperref}

\def\BibTeX{{\rm B\kern-.05em{\sc i\kern-.025em b}\kern-.08em
    T\kern-.1667em\lower.7ex\hbox{E}\kern-.125emX}}
\begin{document}

\makeatletter
    \newcommand{\linebreakand}{
      \end{@IEEEauthorhalign}
      \hfill\mbox{}\par
      \mbox{}\hfill\begin{@IEEEauthorhalign}
    }
\makeatother

\newcommand{\Unit}[1]{\,\mathrm{#1}}
\newcommand{\phil}{\color{blue}}
\newcommand{\daniel}{\color{green}}
\newcommand{\steffen}{\color{cyan}}
\newcommand{\marc}{\color{orange}}
\newcommand{\volkmar}{\color{magenta}}

\title{Learning Control Policies for Variable Objectives from Offline Data
}

\author{\IEEEauthorblockN{Marc Weber}
\IEEEauthorblockA{\textit{Siemens AG, Technology} \\
Munich, Germany \\
marc-weber@siemens.com
}
\and
\IEEEauthorblockN{Phillip Swazinna}
\IEEEauthorblockA{\textit{Siemens AG, Technology} \\
Munich, Germany \\
phillip.swazinna@siemens.com
}
\and
\IEEEauthorblockN{Daniel Hein}
\IEEEauthorblockA{\textit{Siemens AG, Technology} \\
Munich, Germany \\
hein.daniel@siemens.com
}
\linebreakand 
\IEEEauthorblockN{Steffen Udluft}
\IEEEauthorblockA{\textit{Siemens AG, Technology} \\
Munich, Germany \\
steffen.udluft@siemens.com
}
\and
\IEEEauthorblockN{Volkmar Sterzing}
\IEEEauthorblockA{\textit{Siemens AG, Technology} \\
Munich, Germany \\
volkmar.sterzing@siemens.com
}
}

\maketitle

\begin{abstract}
Offline reinforcement learning provides a viable approach to obtain advanced control strategies for dynamical systems, in particular when direct interaction with the environment is not available. In this paper, we introduce a conceptual extension for model-based policy search methods, called variable objective policy (VOP). With this approach, policies are trained to generalize efficiently over a variety of objectives, which parameterize the reward function. We demonstrate that by altering the objectives passed as input to the policy, users gain the freedom to adjust its behavior or re-balance optimization targets at runtime, without need for collecting additional observation batches or re-training.
\end{abstract}


\section{Introduction}
As reinforcement learning (RL) is becoming increasingly widespread in real-world applications, the expectations of practitioners towards higher levels of flexibility and interactivity in control policies are rising, too. Allowing users to maintain a certain level of control over the  policy in operation, such as balancing between competing performance aspects, can thus improve the general acceptance level and applicability of RL methods in industrial domains.

An issue rarely addressed by prior works is that the initially chosen reward function might turn out to not exactly reflect the user's preferences in practical applications. Moreover, these preferences may change at runtime: For example, as an operator of a gas turbine power plant, it may be desirable to maximize the power output during times of high demand, but otherwise minimize consumption, component degradation, or emissions, with any prioritization of these targets being a possible choice of the operator.

Another key requirement for many real-world problems is the ability of an RL algorithm not only to learn by immediate interaction with the environment (\mbox{``online''} learning) but also to find (near)-optimal policies from an immutable batch of observations generated by some prior control strategy (\mbox{``offline''} learning). The latter capability is of particular interest in real-world systems for which excessive exploration phases during training are either prohibited or very expensive (e.g., in power generation, traffic control, or process industries).

We present an offline RL algorithm which allows the user to adjust the behavior of the policy during runtime, in order to meet individual preferences. The preferences are encoded as parameters of the reward function and termed ``objectives''. They are also passed as input to the trained policy, so that it is able to condition its behavior on the objectives that it currently needs to fulfill.
Consequently, the policy needs to be trained to optimize performance not only for one possible objective but all conceivable ones. We will refer to the approach as \textit{variable objective policy} (VOP) in the further course of this paper. 

Before we provide a detailed description of the VOP architecture and training procedure in the main part of our report, we briefly discuss context and related work in the following section. Performance results of our approach are eventually presented and compared to a state-of-the-art method based on a generalized cart-pole upswing and balancing benchmark. We complete our evaluation of the VOP method using the industrial benchmark~\cite{hein_ib}, which is designed to emulate complex dynamical systems, such as industrial processes or turbine control.

\section{Related work}
\label{section:background}
\paragraph{Data-efficient RL}
The need for data-efficient RL in continuous state spaces has led to methods that both store and reuse all observational data and use well generalizing function approximators, such as FQI \cite{ernst2005tree} with ensembles of regression trees, and NFQ \cite{10.1007/11564096_32} and later DQN \cite{mnih2013playing} with neural networks (NNs).
\paragraph{Batch RL} These methods are batch methods and can in principle also be used for offline RL, i.e., the problem when only a fixed dataset is available and a policy is to be created without further interaction with the environment \cite{pmlr-v97-fujimoto19a,levine2020offline}. Extensions are on the one hand the use of policies for continuous actions e.g., PGNRR \cite{10.1007/978-3-540-74690-4_12}, RCNN \cite{4220827}, and PILCO \cite{Deisenroth2011c}, whereby the last two are model-based.
\paragraph{Offline RL}
On the other hand, for offline RL, work has been done to create reliable policies even in the case of incomplete coverage of the state-action space. This is partly already the case for PILCO \cite{Deisenroth2011c} and policy search using Bayesian neural networks  \cite{depeweg2017learning}, where models are used, that incorporate uncertainty. As a central motif it was then addressed in e.g., BCQ \cite{pmlr-v97-fujimoto19a}, BEAR \cite{NEURIPS2019_c2073ffa}, 
BRAC \cite{wu2019behavior}, and CQL \cite{NEURIPS2020_0d2b2061}.
Especially in offline RL, model-based methods became popular, like MOPO \cite{NEURIPS2020_a322852c}, MOReL \cite{NEURIPS2020_f7efa4f8}, MOOSE \cite{Swazinna2021}, COMBO \cite{NEURIPS2021_f29a1797}, and RAMBO-RL \cite{rigter2022ramborl}. Whereby there are methods with model and Q-function (MOPO, MOReL, COMBO, RAMBO-RL) and those with model and without Q-function \cite{4220827,Deisenroth2011c,depeweg2017learning,Swazinna2021}). An overview is given in \cite{SWAZINNA202219}.
\paragraph{Variable objectives}
The present work builds on MOOSE \cite{Swazinna2021}, a model-based method without Q-function, which models stochasticity of dynamics and uncertainty of the estimation of dynamics by an ensemble of neural networks.
The key element is to provide offline RL policies with runtime flexibility so that the user can adjust the policy for variable targets at runtime.
Control over the behavior of the policy at runtime was also addressed in LION \cite{swazinna2023userinteractive}, but there for the tradeoff between return optimality with respect to the estimated MDP and regularization terms, similar to \cite{abdolmaleki2021multiobjective} and \cite{hong2023confidenceconditioned}.
Here we are concerned with variable objectives, which are not determined by regularization terms, but exclusively by the reward function. These variable objectives can be different target states as in \cite{Deisenroth2011d,bischoff2013learning}, or the trade-off of different reward components. The approach allows general transformations of the reward function. The parameters of the reward function are used analogously to the state as input of the policy.
This scheme was used in a similar form in \cite{PIANOSI201110579,5874921,10.2166/hydro.2013.169,pmlr-v37-schaul15,PARISI20173,pong2018temporalICLR,plappert2018multigoal,ijcai2022p0770} where the parameters are instead input to the Q-function. In contrast to these works, our approach is a dedicated offline RL procedure that is purely model-based, does not use a Q-function, and uses an explicit representation of the policy for continuous actions.

\begin{figure*}[t]
     \centering
     \begin{subfigure}[t]{0.49\textwidth}
         \centering
         \includegraphics[width=\columnwidth]{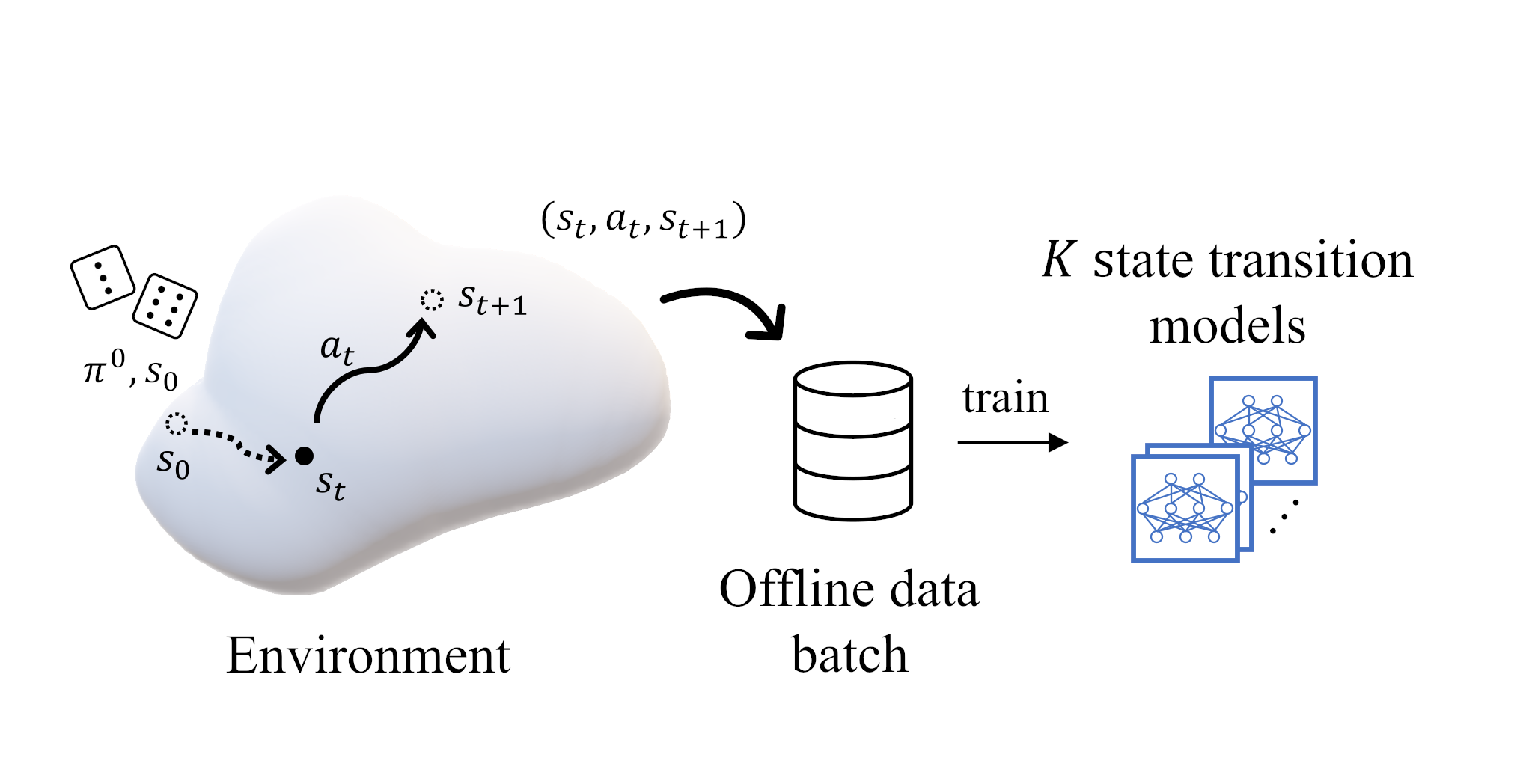}
         \caption{The offline data batch contains state-action-next-state tuples collected from the environment by applying the generating policy $\pi_0$, starting from random initial states. Based on this information an ensemble of NN state-transition models is trained and further used for training the VOP as shown in (b).}
         \label{fig:offline_rl}
     \end{subfigure}
     \hfill
     \begin{subfigure}[t]{0.49\textwidth}
         \centering
         \includegraphics[width=\columnwidth]{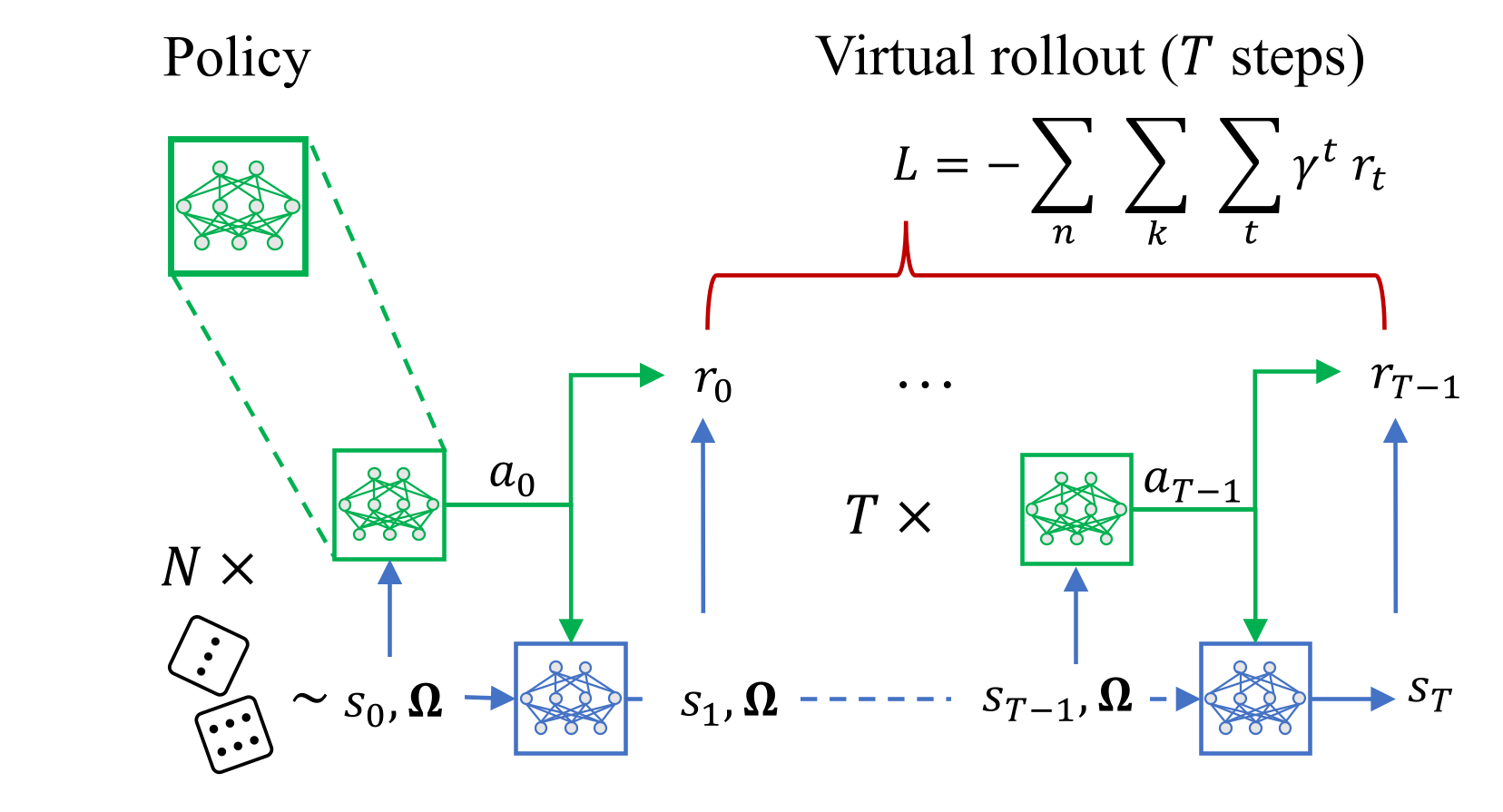}
         \caption{The weights of an NN policy model are trained to minimize the total loss $L$, aggregated over $N$ virtual policy rollouts, using the fixed-weight NN transition model ensemble trained in (a). Starting positions $s_0$ and objective parameters $\Omega$ are drawn from a random distribution at the beginning of the policy training.}
         \label{fig:virtual_rollout}
     \end{subfigure}
     \caption{Illustration of the data and model workflow architecture (a), and the structure of the training algorithm for the VOP (b).}
     \label{fig:vop_concept}
\end{figure*}

\section{Variable objective policy algorithm}
\label{sec:algorithm}
\paragraph{Setup and assumptions} Our proposed offline policy search algorithm is built upon the common assumption that a pre-defined data batch $\mathcal{D}$ of observation tuples $\{(s_t, a_t, s_{t+1})|t=0,\ldots,T-1\}$, obtained by applying some generating policy $\pi_0$ in a given environment for a total of $T$ time steps, provides the only source of information about the environment. Therein, $s_t$ denotes the (observable) Markov state of the system at time $t$, and $a_t$ the action applied by the initial policy $\pi_0$ to perform a transition into the next state $s_{t+1}$. In distinction to the standard problem definition, we do not expect the reward $r_t$, resulting from each state-transition, to be explicitly included in the obtained observation batch. With regard to our intended study of variable objectives, we instead make the assumption that the reward obtained at any time step $t$ can be described by the user as a differentiable function $r$ of state $s_{t}$, action $a_t$, next state $s_{t+1}$, and a set of objective parameters $\Omega$.
\begin{equation}
r_t = r(s_t, a_t, s_{t+1}, \Omega).
\end{equation}
Note that if $\Omega$ are used to explicitly represent target positions within the state space, this formulation is similar to the one assumed in target-oriented
RL~\cite{Deisenroth2011d,bischoff2013learning,pong2018temporalICLR}. Regarding our approach, we drop such limitation, allowing  $\Omega$ to form arbitrary differentiable alterations of the reward signal.

The eventual learning task for our RL~algorithm is to find a policy $\pi^*$ which maximizes the sum of discounted rewards $r_t$,
\begin{equation}
R = \sum_{t=0}^{T-1} \gamma^t r_t,
\end{equation}
when consecutively applied over a time period of $T$ steps.

In offline~RL, the ability to learn such an improved policy generally depends on the choice of the data generating policy $\pi_0$, and there are many studies related to describing the impact for various learning task conditions~\cite{RL_Unplugged,D4RL,NEORL,SWAZINNA202219}. Under almost all circumstances, the initial policy must feature some degree of exploratory behavior in order to generate sufficient coverage of the state-action space. In our present work, we do not put emphasis on the selection of $\pi_0$ but rather consider observation batches generated by a policy which draws from uniformly distributed random actions within the defined boundaries of the given environment. Not only does this reflect a comparable baseline for many common benchmarks, but also allows us to better compare policy performance attributes between different RL~methods, separately from their dependence on particular attributes of the data generation process.

\paragraph{Algorithm} While there has been made noticeable progress on model-free, $\mathcal{Q}$-function based algorithms for off-line learning tasks in the recent past \cite{pmlr-v97-fujimoto19a,wu2019behavior,NEURIPS2019_c2073ffa,NEURIPS2020_0d2b2061},
we follow the concept of model-based policy search in our present work. As previously outlined in \cite{Swazinna2021}, virtual trajectories (also referred to as rollouts) are performed using surrogate dynamics models to train the policy. The dynamics models are trained on one-step state-transitions from the initial dataset $\mathcal{D}$, and often yield a more stable and accurate estimation of the expected return $R$ compared to model-free methods, as long as the regression error of those models can be kept sufficiently small.

Similarly to \cite{Swazinna2021}, we first train an ensemble of $K$ feed-forward neural networks $\{\hat{T}^k_\phi, k=0,\ldots,K-1\}$, using the step-wise observations contained in $\mathcal{D}$ (Fig.~\ref{fig:offline_rl}). These models simulate the real transition dynamics function $T^*$, and each aims to predict the next environment state $\hat{s}_{t+1}$ given the prior state $s_t$ and applied action $a_t$. We normalize inputs and outputs of the networks by their respective means ($\mu^s$) and standard deviations ($\sigma^s)$ and transform the prediction targets into normalized differences ($\Delta s')$ between consecutive states. Ensemble members are then trained to minimize the mean squared error of these one-step differences by means of gradient descent:
\begin{equation}
    L(\phi^k) = \mathbb{E}_{s,a,s'\sim\mathcal{D}} \left(\hat{T}^k_\phi\left(\frac{s-\mu^{s}}{\sigma^{s}},a\right) - \frac{\Delta s' - \mu^{\Delta s}}{\sigma^{\Delta s}}\right)^2 .
\end{equation}

The ensemble is then used to construct virtual rollouts, which represent a learned version of the real Markov decision process (MDP). The experience to train the policy will thus be entirely generated by the dynamics models instead of the true MDP, enabling policy learning without environment interaction. To predict next states $\hat{s}'$ 
based on some state $s$ and action $a$, we construct the ensemble function $\hat{T}$ as the average prediction of the ensemble members:
\begin{equation}
    \hat{T}(s,a) = \frac{1}{K}\sum_{k=0}^{K-1} \hat{T}^k_\phi(s,a).
\end{equation}
The virtual rollouts then all start in a state $s_0$ sampled from the initially collected dataset $\mathcal{D}$, and continue by alternatingly evaluating the policy for the current state to obtain a new action and then estimating the next state by feeding the state and action into the ensemble. To this end, a policy model $\pi_\psi$ provides the next action $\hat{a}_t$ which is passed on to the dynamics models $\hat{T}$ for the next state estimation (Fig.~\ref{fig:virtual_rollout}). 

Since the behavior of the policy should be altered based on the objectives provided, we also pass objective parameters $\Omega$, each sampled from a random uniform distribution $p(\Omega)$ over an interval of possible values, as additional input to the policy. Virtual trajectories $\mathcal{T}(s_0, \Omega)$ for some policy $\pi$ and a given transition ensemble $\hat{T}$ are then defined as:
\begin{equation}
    \mathcal{T}^\pi(s_0,\Omega) = \{(s_0, \hat{a}_0), (\hat{s}_1, \hat{a}_1), \ldots, (\hat{s}_{H-1}, \hat{a}_{H-1}), \hat{s}_H\},
\end{equation}

where $s_0 \sim \mathcal{D}$, $\hat{a}_t = \pi_\psi(\hat{s}_t, \Omega)$, and $\hat{s}_{t+1} = \hat{T}(\hat{s}_t, \hat{a}_t)$. Note, that in our setup $\Omega$ samples remain fixed over the duration of each virtual trajectory.

\paragraph{Rewards} Next, we incorporate the reward signals collected along the trajectories into the algorithm by evaluating the reward function $r(s,a,s',\Omega)$ for every corresponding state-action-next-state tuple and current choice of objectives $\Omega$. By averaging over all virtual trajectories and the sum of (discounted) rewards therein, we finally obtain the expectation value of the policy's achieved return $\hat{R}$:
\begin{equation}
    \mathbb{E}[\hat{R}] = \mathbb{E}_{s_0 \sim \mathcal{D}, \Omega \sim p(\Omega), \tau = \mathcal{T}^\pi(s_0,\Omega)} \left[\sum_t \gamma^t r(\tau_t, \Omega)\right],
\end{equation}
where $\tau_t=(s_t, a_t, s_{t+1})$.

Once we have estimated the return over a batch of trajectories resulting from different starting states and objectives, we can use it directly to optimize the policy parameters $\psi$ through back-propagation if we assume the reward function $r(\cdot)$ to be differentiable. The loss function we seek to minimize by gradient descent to train the VOP is thus the negative estimated return:
\begin{equation}
    L(\psi) = - \mathbb{E}[\hat{R}].
\end{equation}

Noteworthy about the model-based formulation using the virtual rollouts $\mathcal{T}(\hat{s}_0, \Omega)$ is that the transition models $\hat{T}^k$ can be trained entirely in isolation and be evaluated using supervised learning methods. During policy training, the weights of the dynamics models can then be frozen since their predictions do not depend on the specified objective parameters $\Omega$. In contrast, if a model-free approach would have been chosen, a value function would need to incorporate them when predicting the value of a state (or state-action pair), since the value depends on the currently desired objectives. This means that two intertwined components need to be trained jointly, which prior work has hypothesized to lead to instabilities in the learning process \cite{swazinna2023userinteractive}.

\section{Experimental results}
\subsection{Cart-pole upswing and balancing}
We begin our evaluation of the VOP algorithm on an extended version of the classic cart-pole upswing benchmark. A schematic drawing of the simulated environment is shown in Fig.~\ref{fig:cartpole_setup}. The baseline task is to swing-up a pole and keep it balanced, with the pole connected to a moving cart that can be accelerated left and right along a horizontal $x$-axis by exerting a force $F_a$, which is proportional to an action signal $a_t$. Our environment is implemented as an extension of the classic openAI cart-pole simulator~\cite{openai_gym}, in which the pole is allowed to move within the entire angular space, defined between $-\pi$ and $\pi$. 
By including the velocities $\dot{x}$ and $\dot{\theta}$ of the cart and pole, respectively, the observation tuple $(x, \theta_t, \dot{x}_t, \dot{\theta}_t)$ describes the Markov state of the system.

\begin{figure}[t!]
\centering
\includegraphics[width=0.8\columnwidth]{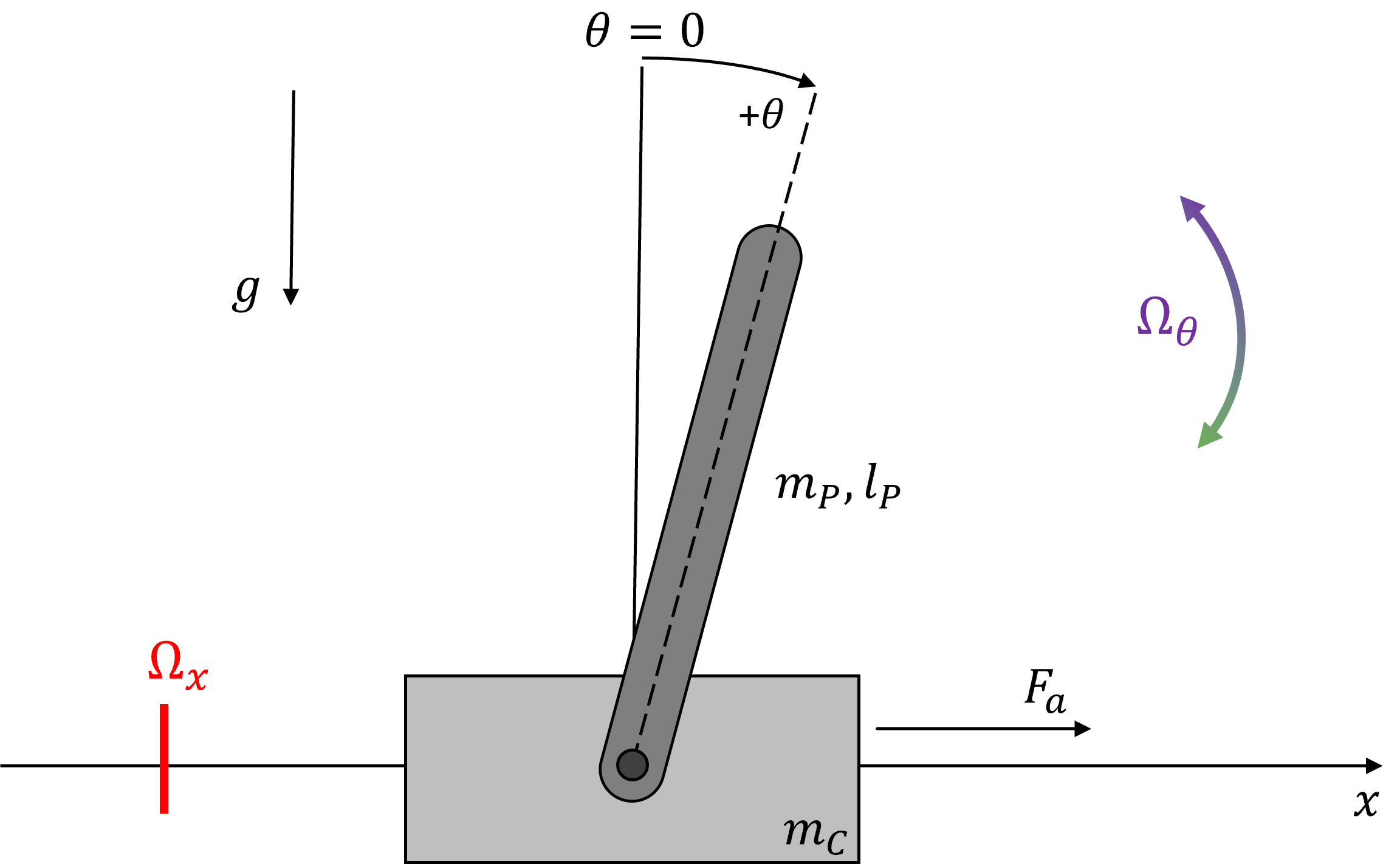}
\caption{Schematic drawing of the cart-pole upswing and balancing setup with our VOP extensions highlighted in red and purple. The two variable objective parameters, $\Omega_x$ and $\Omega_{\theta}$, encode continuous target settings in $x$-axis and upswing behavior, respectively.}
\label{fig:cartpole_setup}
\end{figure}

\paragraph{Rewards}
We extend the usual reward definition, which encompasses one term for the cart position $x \in [-2.5\Unit{m}, 2.5\Unit{m}]$ along the horizontal axis and one for the pole alignment $\theta \in [-\pi, \pi)$, by two objective parameters $\Omega_{x}$ and $\Omega_{\theta}$ in the following way:
\begin{equation}
    r(x, \theta; \Omega_x, \Omega_\theta) = - | x - \Omega_x | - \Omega_\theta \, |\theta| - \tilde{r}(\dot{\theta})
\label{eq:cartpole_reward}
\end{equation}
Thus, $\Omega_x$ defines the targeted $x$-position and $\Omega_\theta$ a weight factor describing the pole's balancing behavior (larger values of $\Omega_\theta$ imply stronger emphasis on keeping the pole in upright position, while lower values relax this penalty). The remaining term, $\tilde{r}(\dot{\theta})$, only affects small alignment angles $\theta$ and is introduced to improve smoothness in the balancing behavior of successful policies:
\begin{equation}
    \tilde{r}(\dot{\theta}) = \begin{cases} |\dot{\theta}/ 2 |, & |\theta| < 0.4 \\ 
    0, & \mathrm{otherwise}.
    \end{cases}
\end{equation}

\paragraph{Training} The VOP~algorithm is subsequently trained using an offline dataset consisting of $1000$ episodes, each limited to a maximum length of $250$ time steps, generated by an initial policy $\pi_0$. This policy is state-independent and generates continuous, uniformly distributed random actions $a_t \in [-2, 2]$, which are linearly scaled to a one-dimensional force $F_a \in [-20\Unit{N},20\Unit{N}]$ applied to the moving cart in the described simulator. Each episode starts with different initial values for $x$, $\theta$ and the respective objective parameters, $\Omega_x$ and $\Omega_{\theta}$---all drawn from random uniform distributions within fixed box limits (for $x$ and $\theta$ those box regions correspond to their allowed value ranges; the objective values for $\Omega_x$ and $\Omega_{\theta}$ are in $[-2,2]$ and $[0,4]$, respectively, and are kept fixed until the end of the respective episode). Note that our fixed observation sample of $1000$ randomly explored episodes represents only a fraction of the vast number of possible initial state-objective combinations and, hence, requires any successful policy to generalize beyond the states and objectives seen in training. Regarding the ensemble size for the state-transition models trained on the generated observation batch we choose $M\!=\!8$. Each member is represented by a fully connected, 2-layer feed-forward NN with a capacity of $20$ neurons per layer and linear rectifier activations. The policy is described by the same model architecture but uses a reduced number of $10$ neurons per layer.  

\begin{figure}[t!]
\includegraphics[width=1.0\columnwidth]{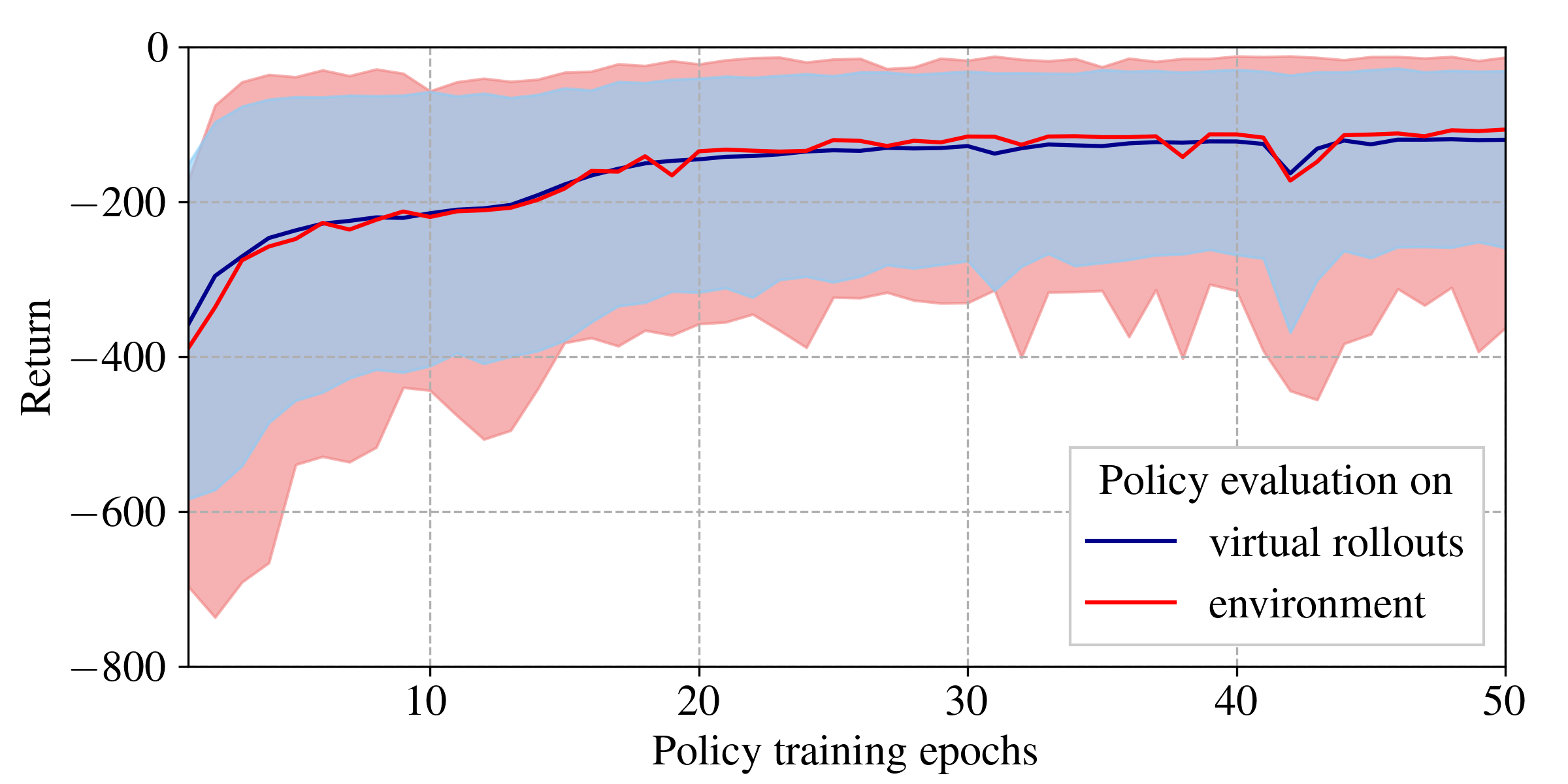}
\caption{Return improvement as function of VOP training epochs for the cart-pole upswing benchmark. Shaded areas describe full width of distributions resulting from random sampling of initial conditions and objectives.}
\label{fig:cartpole_learning_curve}
\end{figure}

Experimentally, we figure that a reasonable minimum time scale in order to bring the pole in upright position (even when starting close to the downward angular position at $\theta\!=\!\pm\pi$) for an average remote $\Omega_x$ target position is about $65$ time steps. During policy training, this value is chosen as the virtual rollout horizon $T$. We observe a deteriorating policy training performance for too large values of $T$ ($>80$), mainly because the aggregated dynamics modelling error for such long horizons starts becoming too large.

\begin{figure}[t!]
\centering
\includegraphics[width=1.0\columnwidth]{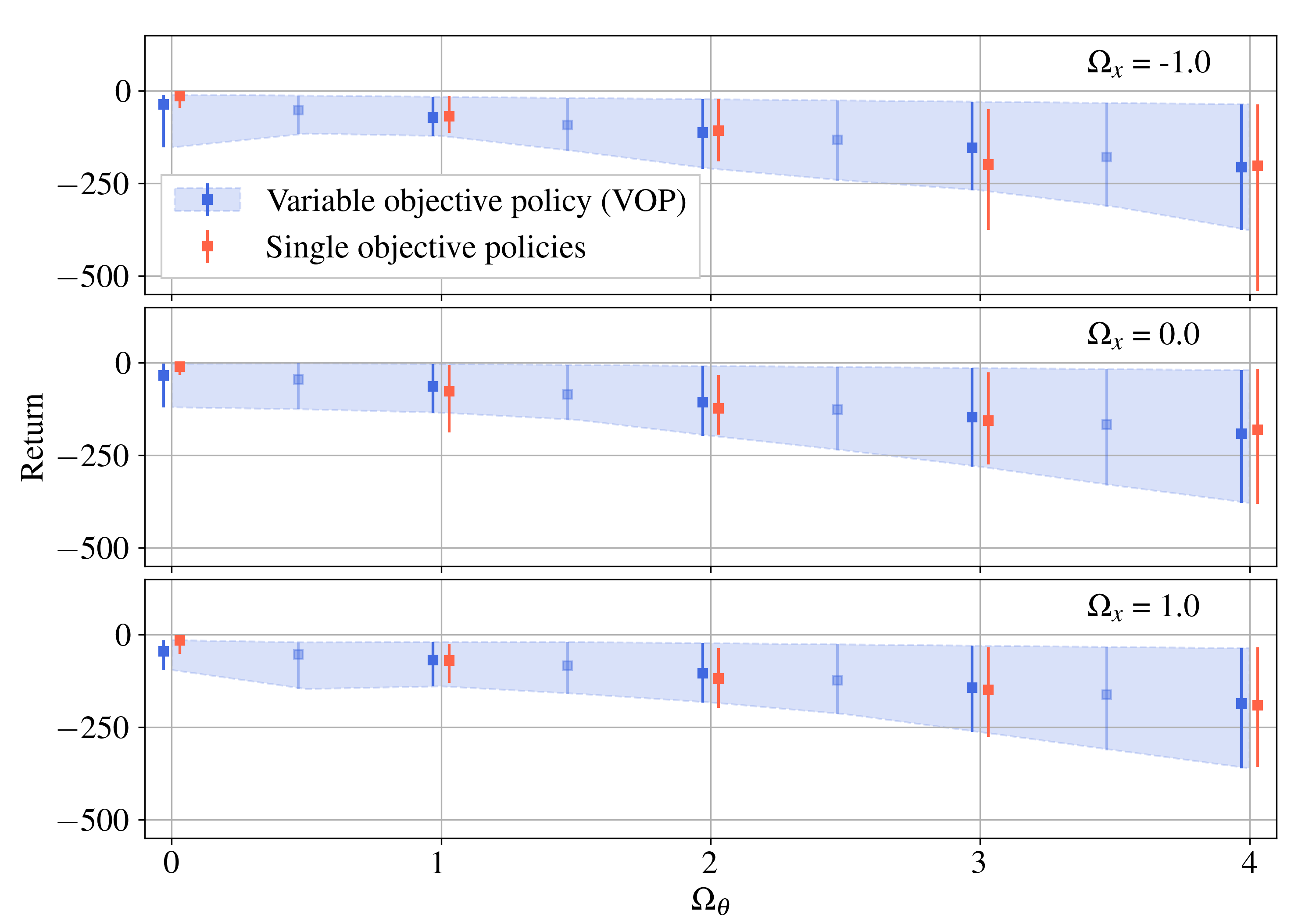}
\caption{Comparison of returns achieved after $250$ simulated steps by the trained VOP (blue) at selected objectives, $\Omega_x$ and $\Omega_\theta$, and individual policies trained exclusively for each objective value pair (red). Error bars indicate the full range of returns for a fixed set of $100$ random start positions of the cart and the pole.}
\label{fig:compare_vop_vs_single_objectives}
\end{figure}

\begin{table}[t!]
\begin{center}
\resizebox{\columnwidth}{!}{
\begin{tabular}{l c c c}
& \multicolumn{3}{c}{Return after $250$ evaluation steps} \\
\cmidrule{2-4}
Eval. setup & VOP & VOP & CQL \\
$(\Omega_x, \Omega_{\theta})$ & Virt. rollout & Env & Env \\ 
\midrule
\begin{minipage}{0.15\columnwidth}\vspace{4pt}$(\text{-}1, 0)$\vspace{4pt} \end{minipage} & $-34\pm1$ & $-44\pm4$ &  $-57\pm5$ \\ \hline
\begin{minipage}{0.15\columnwidth}\vspace{4pt}$(\text{-}1, 3)$\vspace{4pt} \end{minipage} & $-162\pm3$ & $-166\pm6$ & $-226\pm12$  \\ \hline
\begin{minipage}{0.15\columnwidth}\vspace{4pt}$(0, 1)$\vspace{4pt} \end{minipage} & $-74\pm2$ & $-74\pm4$ & $-523\pm16$   \\ \hline
\begin{minipage}{0.15\columnwidth}\vspace{4pt}$(1, 2)$\vspace{4pt} \end{minipage} & $-123\pm2$ & $-121\pm6$ & $-175\pm4$  \\ \hline
\begin{minipage}{0.15\columnwidth}\vspace{4pt}$\sim p(\Omega)$\vspace{4pt}\end{minipage} & $-117\pm2$ & $-121\pm5$ & n.a.  \\
\bottomrule
\end{tabular}}
\end{center}
\caption{Comparison of cart-pole benchmark results achieved by our VOP and the model-free offline RL algorithm CQL~\cite{NEURIPS2020_0d2b2061} for different policy evaluation settings. We perform $10$ trials with varying random seeds, each aggregating performance across the same 100 randomly sampled starting states.}
\label{tab:upswing_results}
\end{table}

In Fig.~\ref{fig:cartpole_learning_curve}, we show the improvement of the VOP performance during training, measured by the return as a function of policy network training epochs and averaged over the variable objective parameters $\Omega_x$ and $\Omega_\theta$. Therein, as well as in Tab.~\ref{tab:upswing_results}, we compare the distribution of returns according to the model rollouts (upon which the policy is trained to improve) with the distribution resulting from the episodes in the real upswing environment. Both distributions are found to be consistent during the entire training process, confirming that the return estimates based on the virtual rollout models provide a sound representation of the corresponding metric in the actual environment. After about $40$ training epochs, the measured performance reaches a plateau, at which the learning process is stopped. 

The VOP has learned to generalize well over an arbitrary number of objectives and, thus, is adaptive towards different user demands at execution time. It is further demonstrated in Fig.~\ref{fig:compare_vop_vs_single_objectives} that performance for specific values of the objective parameters is on par with the result observed for individual policies, trained only on these values. This is an important finding since obvious benefits in terms of flexibility and resource-efficiency are not compromised by a general drop in performance. 

\paragraph{Performance}
The performance of the VOP for selected test objectives is summarized in Tab.~\ref{tab:upswing_results} and compared with the results of CQL~\cite{NEURIPS2020_0d2b2061}, a state-of-the-art model-free offline RL algorithm. Benchmark results for CQL are generated by evaluating a set of individual policies, each one trained on the same observation batch as the VOP but using a fixed setup of ($\Omega_x$, $\Omega_\theta$) values. We report mean and standard error of the evaluation returns after averaging over a set of $100$ episodes with identical start conditions and repeat each experiment $10$ times with varying random seeds of the respective algorithms. CQL policies are tested at the end of every training epoch, and the one with maximum return is taken into account---although some learning curves exhibit a slight decline after reaching an intermediate plateau around $\approx\!50$ epochs. We observe that our VOP shows significantly better performance in virtually every scenario. Optimal CQL policies only manage to swing up and balance the pole for $\Omega_{\theta}\geq2$ but fail to simultaneously reach the desired target position $\Omega_x$ in the required 250 steps, which qualitatively explains the performance discrepancy. 

\paragraph{Illustrative example} In Fig.~\ref{fig:eval_upswing_modes} we show an illustrative example of how different objectives impact the qualitative behavior of the trained VOP at runtime. In the beginning of the experiment, the chosen objectives direct the policy to swing-up the pole at $\Omega_x\!=\!-1\,\mathrm{m}$ before a sudden change to $\Omega_x\!=\!+1\,\mathrm{m}$ is injected at time step $t\!=\!150$. One can observe how the VOP reacts in different ways depending on the value of the other objective parameter, $\Omega_\theta$. In case of a higher weight ($\Omega_\theta\!=\!3$, purple curve), the policy maintains the pole in almost upright position while traversing to the new position. In the other case ($\Omega_\theta\!=\!1$, green curve), the policy applies to a different strategy and lets the pole swing through during the translatory movement, thus reaching the new $x$-position faster.\\
This illustrates the utility of VOP: If the user's preference is that the policy continues to balance even when the position changes, a non-VOP policy would not meet the user's expectations unless $\Omega_\theta$ was chosen sufficiently high for the training. Thus, without VOP, there is the problem that it is not trivial to design the reward function in such a way that the desired behavior is realized. By using the VOP, the user can adapt the behavior to one's preferences at runtime and also has the freedom to decide whether the position change should take place with the pole balanced and therefore slow, or with the pole swinging through and therefore fast.

\begin{figure}[t]
\begin{subfigure}{0.49\columnwidth}
\centering
\frame{\includegraphics[width=0.7\columnwidth]{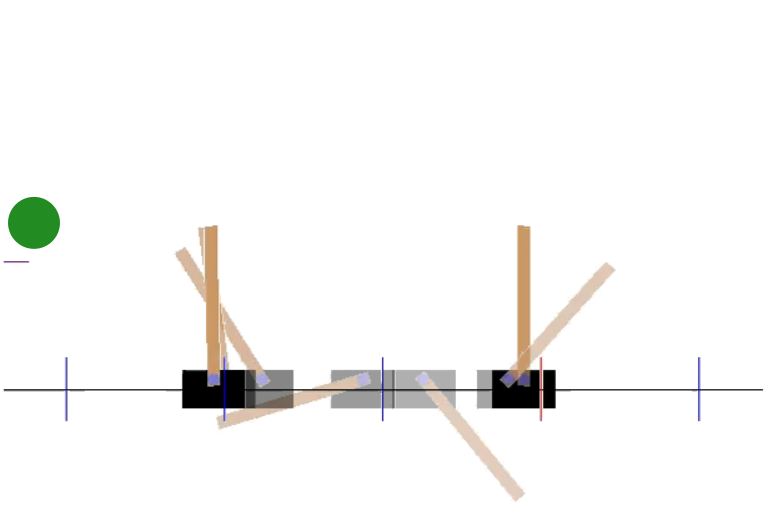}}
\caption*{$\Omega_{\theta} = 1$}
\end{subfigure}
\begin{subfigure}{0.49\columnwidth}
\centering
\frame{\includegraphics[width=0.7\columnwidth]{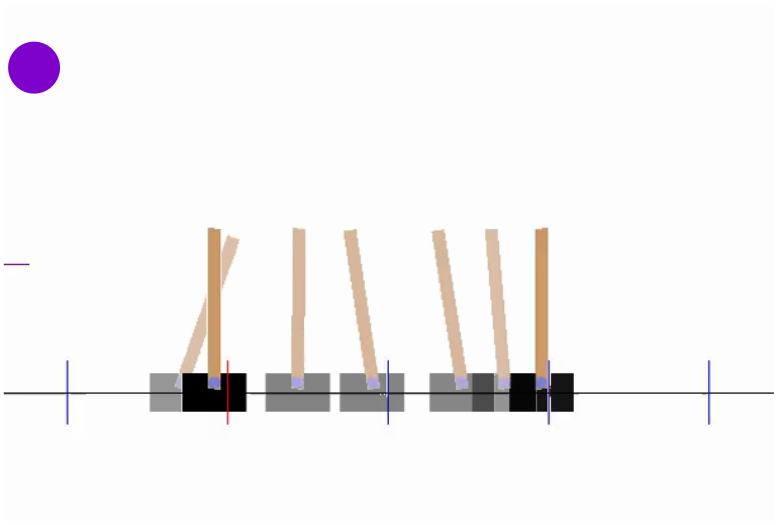}}
\caption*{$\Omega_{\theta} = 3$}
\end{subfigure}
\begin{subfigure}{1.0\columnwidth}
\includegraphics[width=1.0\columnwidth]{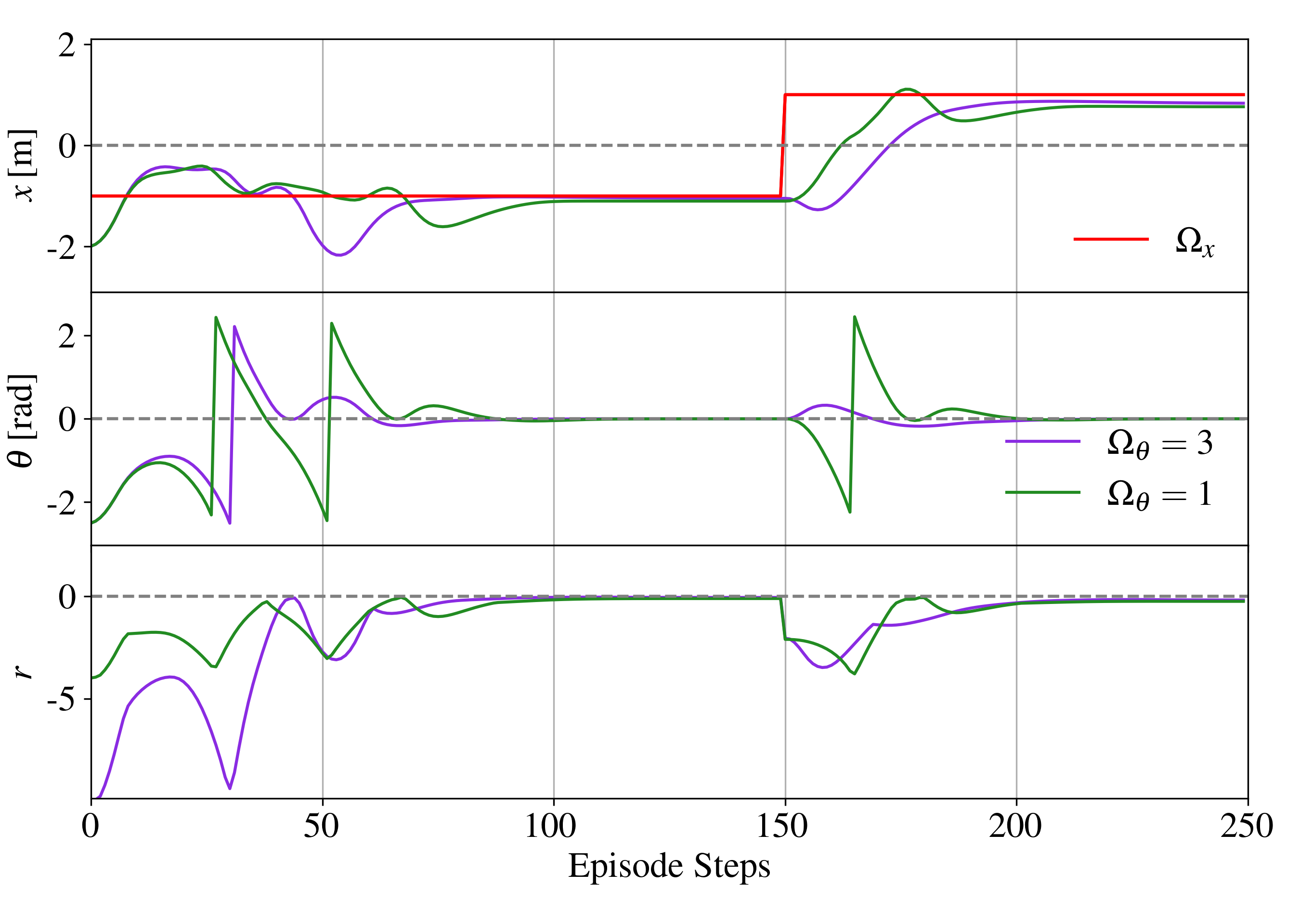}
\end{subfigure}
\caption{Evaluation of our trained VOP for $\Omega_{\theta}\!=\!1$ and $\Omega_{\theta}\!=\!3$. In the beginning the cart is positioned at the leftmost corner with the pole pointing downwards. After $150$~steps the target position is suddenly changed to $\Omega_x\!=\!+1\Unit{m}$. Depending on $\Omega_{\theta}$, the policy either moves to the new position aggressively, thereby dropping the pole in between positions (green), or more carefully, without dropping it (violet).}
\label{fig:eval_upswing_modes}
\end{figure}

\begin{figure}[t]
\includegraphics[width=1.0\columnwidth]{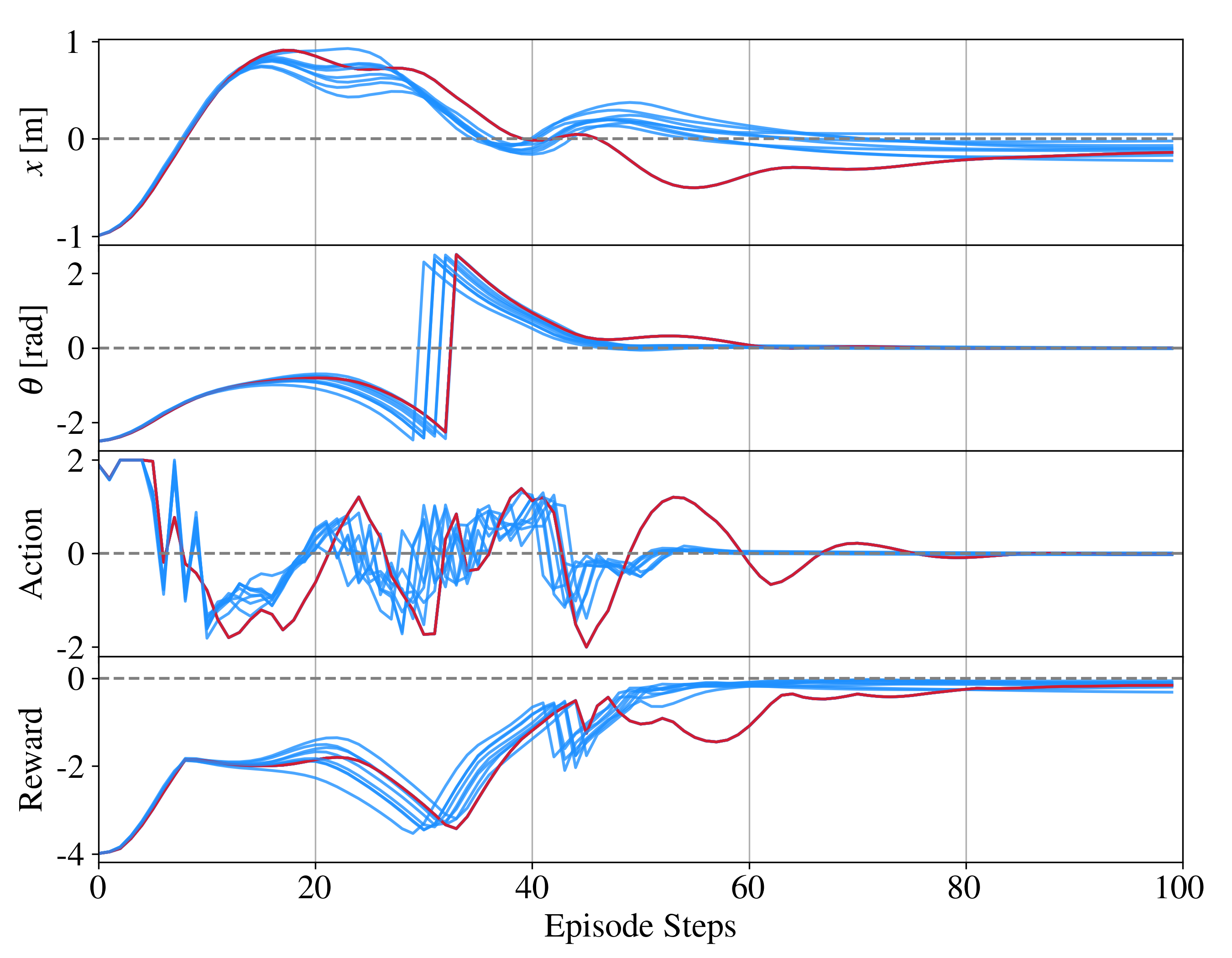}
\caption{
Transferability of the policy trained on the state-transition model to the actual environment. It can be seen that the policy has not only learned to upswing ($\Omega_\theta\!=\!1$) and balance at the desired position ($\Omega_x\!=\!0$) on the state-transition model (blue lines), but also achieves this result on the actual environment (red line).
}
\label{fig:compare_rollouts}
\end{figure}

\paragraph{Model generalization}
We complete this benchmark analysis by demonstrating another important feature of the model-based VOP approach.
In Fig.~\ref{fig:compare_rollouts}, we compare the virtual policy rollout for a given objective with the application of the same policy in the actual environment. 
This policy, which has been trained via the rollouts in the state-transition model, has not only learned to upswing and balance at the desired position when applied to that state-transition model (blue lines), but also achieves this result on the actual environment (red line). 
Naturally, even small imperfections of the state-transition model accumulate, leading to different actions performed during the rollouts and during the application to the actual environment.
 Still, the state-transition model is a sufficient estimate of the actual environment's dynamics, in the sense, that the learned policy can be successfully transferred to the actual environment.
Comparing both virtual and actual policy performance can also provide useful insights in case of unsatisfying policy results as it helps to decouple policy-related issues from potential deficiencies of modelling the state-transition dynamics. 
 
\subsection{Industrial benchmark}
We continue our experimental evaluation of the VOP approach using a different type of RL benchmark motivated by real-world challenges in industrial control applications such as complex combustion engines, referred to as \textit{industrial benchmark} (IB), and described in detail by \cite{hein_ib}. 

In contrast to the previous cart-pole experiment, the dynamics of the IB are characterized by stochastic noise, various time delays and only partially observable states (although agents are allowed to use a history buffer of observations in order to form an approximation of the Markov state). Out of the visible observation signals, consumption ($C$) and fatigue ($F$) are the most important ones to describe the qualitative status of the engine. Their time evolution is indirectly controlled by three steering variables (velocity ($V$), gain ($G$), and shift ($S$)), which make up the remaining part of the observable state space.
Control actions are defined as applying normalized changes within the continuous interval $[-1, 1]$ for each of the three steerings at each time step. A successful policy needs to consider a significant time horizon or otherwise will not be able to arrive at the optimal operation point of long-term stability and relatively low consumption.

\paragraph{Rewards}
By introducing two explicit $\Omega$-dimensions, we generalize the original reward definition at time step $t$ of the IB for our upcoming experiments as follows:
\begin{equation}
    r(C, F, S; \Omega_F, \Omega_S) = -C - \Omega_F F - \Omega_S \sum_{t'=1}^{10} (S_{t-t'} - \mu_S)^2.
\label{eq:ib_reward}
\end{equation}
With this extended formulation, $\Omega_F$ defines the scale of emphasis put on the fatigue level $F$, while $\Omega_S$ weighs a measure equivalent to the stress induced by changing the shift steering $S$ (quantified by computing the variance of $S$ over a rolling time window of $10$ steps in the last term of the above equation). 

\begin{figure*}[t!]
     \centering
     \begin{subfigure}[t]{0.49\textwidth}
         \centering
         \includegraphics[width=\textwidth]{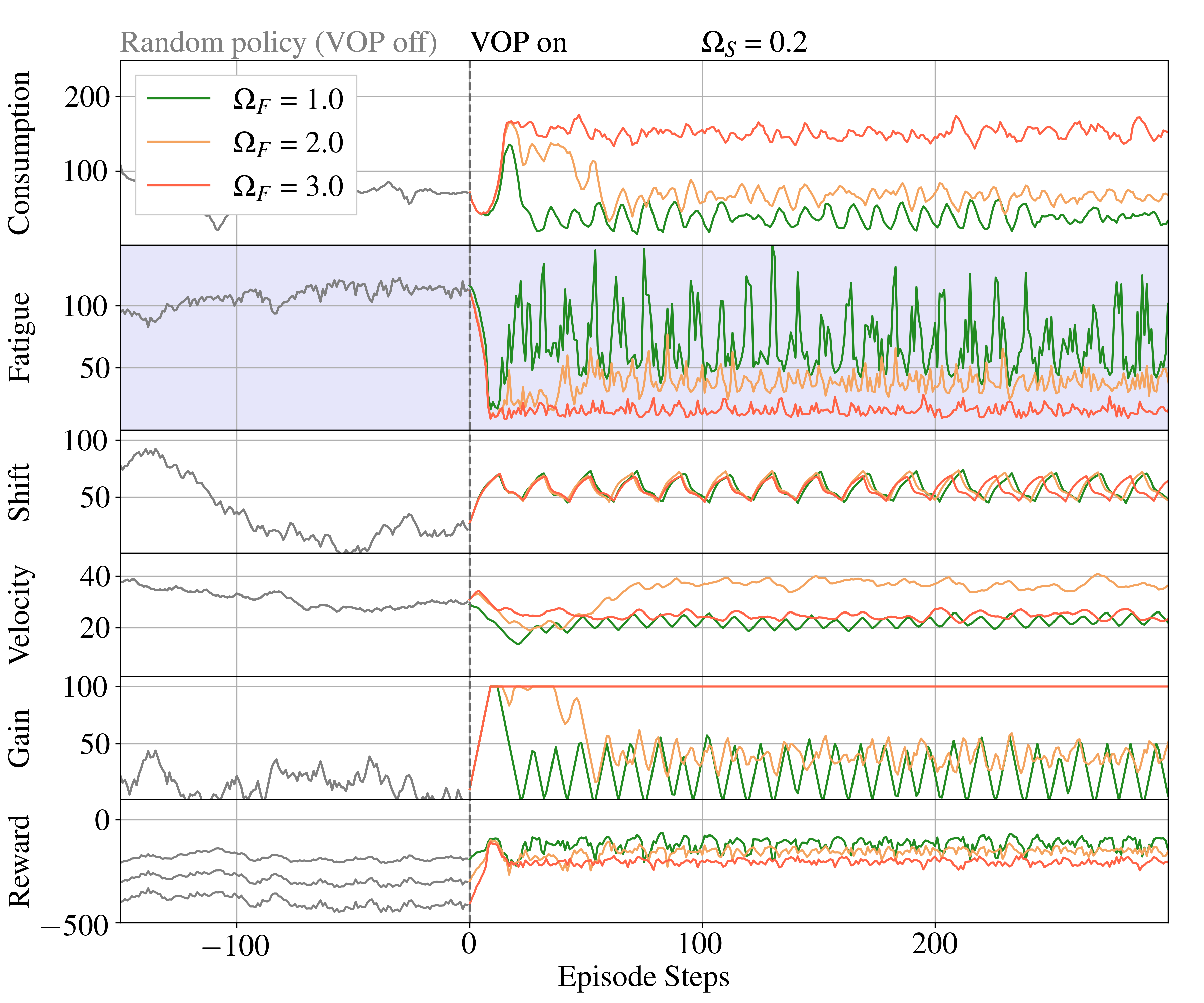}
         \caption{Effect of different fatigue objectives $\Omega_F$ on the qualitative VOP behavior. A stronger factor implies a steeper drop in the fatigue level at the expense of increased consumption.}
         \label{fig:ib_evaluation_omega_f}
     \end{subfigure}
     \hfill
     \begin{subfigure}[t]{0.49\textwidth}
         \centering
         \includegraphics[width=\textwidth]{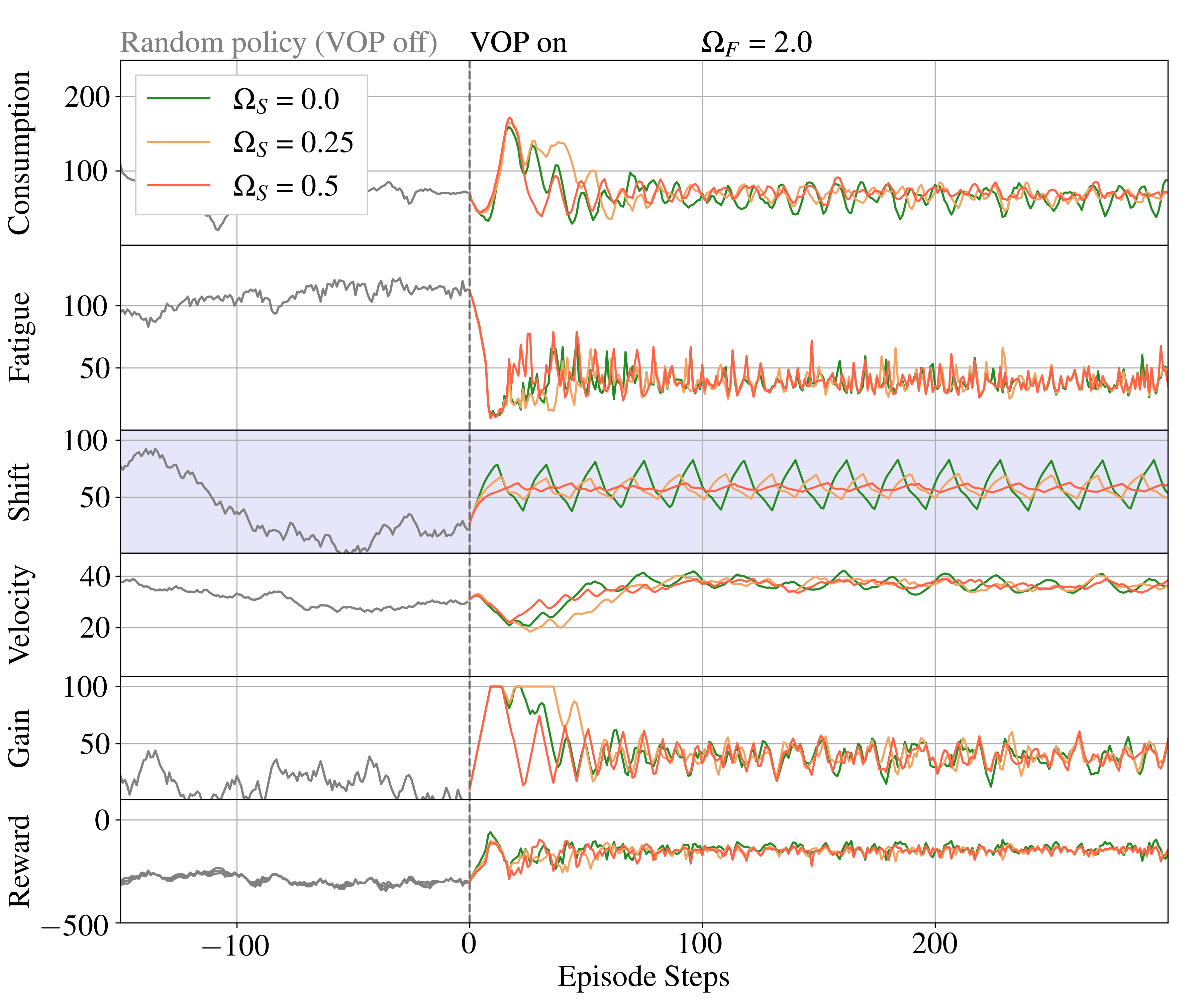}
         \caption{Impact of varying stress attention levels $\Omega_S$ on the shift steering variable. Higher emphasis results in a lower cyclic amplitude caused by the VOP, and thus, corresponds to reduced wear and tear.}
         \label{fig:ib_evaluation_omega_s}
     \end{subfigure}
     \caption{Response of the trained VOP for various objective combinations on the IB benchmark. For all examples the policy is activated after applying $1000$ random action steps on the environment (only $150$ of which are shown within the figures).}
     \label{fig:ib_evaluation}
\end{figure*}

\paragraph{Training}
From the environment an offline data batch, consisting of $200$ episodes with $1000$ steps each, is drawn by applying random actions $\in [-1, 1]$ for each of the three steering control variables.
In addition, the vector-like observations from $10$ consecutive time steps are concatenated to form a rolling observation buffer, which serves as an approximation to the hidden Markov state of the IB, which consists of 20 dimensions in total, i.e., the five mentioned above, the setpoint which we keep fixed at 70, and 14 additional hidden dimensions. The thus expanded states are fed as input to the members of the state-transition models, which are constructed to predict only the feature vector of the next time slice. Then, the resulting observation buffer is shifted accordingly before passing it into the next step in the virtual rollout model. To account for the higher dimensionality of observation space and policy output dimensions in comparison to the previous benchmark, the capacity of the 2-layer feed-forward NN is somewhat enlarged, using layer dimensions of $[40, 20]$ and $[20, 10]$ neurons for the state-transition and policy models, respectively. In each policy training epoch, VOP weights are updated based on the aggregated loss over $N\!=\!2000$ fixed random combinations of initial state and objective parameters, using a rollout horizon of $100$ steps. After $\approx200$ epochs the mean optimization loss $L(\psi)$ has improved from an initial value of $4.3\times10^{4}$ to an asymptotic level of $1.7\times10^{4}$, after which the training process is stopped.

\paragraph{Performance}
For performance testing of the offline-trained VOP on the actual environment, we follow the practice of previous IB analyses~\cite{hein_ib}, in which the mean policy performance is measured after initially applying $\approx \!1000$ random actions. Averaging over $100$ randomized initial conditions, a mean return of $-195\,\pm\,3$ is achieved for the IB standard configuration ($\Omega_F\!=\!3$, $\Omega_S\!=\!0$). This result is on par with other offline RL algorithms that have only been optimized on a single choice of objectives (e.g., refer to \cite{hein2017batch}). This demonstrates the competitiveness of the trained VOP in addition to its capability to generalize over a large range of $\Omega$ targets. 

Qualitative results for various objective combinations, as provided in Fig.~\ref{fig:ib_evaluation}, show in addition that the policy has not only learned to apply the desired cyclic control pattern which allows to achieve stable operation but also to react consistently upon different values along both objective dimensions, $\Omega_F$ and $\Omega_S$. In case of varying fatigue objectives, it is shown in Fig.~\ref{fig:ib_evaluation_omega_f} that the VOP maintains a low fatigue level if $\Omega_F$ is large (at the expense of increased consumption), and vice versa. Regarding its variable response to different setpoints along $\Omega_S$, as demonstrated in Fig.~\ref{fig:ib_evaluation_omega_s}, the policy applies a less aggressive control strategy of the shift steering if $\Omega_S$ is large---without affecting the other performance measure in a negative way. Both situations provide examples of how users can adjust the behavior of the offline-trained VOP at runtime, without need for re-collecting data or re-training the RL policy. 

\section{Conclusion}
In this work, we propose that policies created by offline RL should be flexible at runtime with respect to optimization goals since it facilitates fast adaptation to new settings without retraining, possibly guided by expert users.

We present a novel algorithm for creating variable objective policies (VOPs) and show on the basis of benchmarks that the resulting control strategies effectively generalize across the range of possible objectives.
The experiments show that even under the constraints of offline RL, it is possible to train a policy for a wide range of tasks, parameterizable by continuous objective parameters $\Omega$ in a single training run. In particular, our proposed algorithm does not require that the offline dataset already contains the variety of objectives that will later be integrated into the VOP. This feature enables a computationally and data efficient training procedure, since the sampling process for numerous objectives is now part of the optimization step, rather than part of the data aggregation or model generation phase. Our experiments show that users can adjust the behavior of the trained policy and rebalance the optimization objectives at runtime without the need to collect additional observations or repeat the training. Compared to alternative RL approaches that do not allow for variable objectives, our VOP algorithm is shown to achieve comparable or better results.

\bibliographystyle{IEEEtran}
\bibliography{refs}

\begin{thebibliography}{10}
\providecommand{\url}[1]{#1}
\csname url@samestyle\endcsname
\providecommand{\newblock}{\relax}
\providecommand{\bibinfo}[2]{#2}
\providecommand{\BIBentrySTDinterwordspacing}{\spaceskip=0pt\relax}
\providecommand{\BIBentryALTinterwordstretchfactor}{4}
\providecommand{\BIBentryALTinterwordspacing}{\spaceskip=\fontdimen2\font plus
\BIBentryALTinterwordstretchfactor\fontdimen3\font minus
  \fontdimen4\font\relax}
\providecommand{\BIBforeignlanguage}[2]{{%
\expandafter\ifx\csname l@#1\endcsname\relax
\typeout{** WARNING: IEEEtran.bst: No hyphenation pattern has been}%
\typeout{** loaded for the language `#1'. Using the pattern for}%
\typeout{** the default language instead.}%
\else
\language=\csname l@#1\endcsname
\fi
#2}}
\providecommand{\BIBdecl}{\relax}
\BIBdecl

\bibitem{hein_ib}
D.~Hein, S.~Depeweg, M.~Tokic, S.~Udluft, A.~Hentschel, T.~A. Runkler, and
  V.~Sterzing, ``A benchmark environment motivated by industrial control
  problems,'' in \emph{2017 IEEE Symposium Series on Computational Intelligence
  (SSCI)}, 2017, pp. 1--8.

\bibitem{ernst2005tree}
D.~Ernst, P.~Geurts, and L.~Wehenkel, ``Tree-based batch mode reinforcement
  learning,'' \emph{Journal of Machine Learning Research}, vol.~6, 2005.

\bibitem{10.1007/11564096_32}
M.~Riedmiller, ``Neural fitted {Q} iteration -- first experiences with a data
  efficient neural reinforcement learning method,'' in \emph{Machine Learning:
  ECML 2005}, J.~Gama, R.~Camacho, P.~B. Brazdil, A.~M. Jorge, and L.~Torgo,
  Eds.\hskip 1em plus 0.5em minus 0.4em\relax Berlin, Heidelberg: Springer
  Berlin Heidelberg, 2005, pp. 317--328.

\bibitem{mnih2013playing}
V.~Mnih, K.~Kavukcuoglu, D.~Silver, A.~Graves, I.~Antonoglou, D.~Wierstra, and
  M.~Riedmiller, ``Playing atari with deep reinforcement learning,'' 2013.

\bibitem{pmlr-v97-fujimoto19a}
S.~Fujimoto, D.~Meger, and D.~Precup, ``Off-policy deep reinforcement learning
  without exploration,'' in \emph{Proceedings of the 36th International
  Conference on Machine Learning}, ser. Proceedings of Machine Learning
  Research, K.~Chaudhuri and R.~Salakhutdinov, Eds., vol.~97.\hskip 1em plus
  0.5em minus 0.4em\relax PMLR, 09--15 Jun 2019, pp. 2052--2062.

\bibitem{levine2020offline}
S.~Levine, A.~Kumar, G.~Tucker, and J.~Fu, ``Offline reinforcement learning:
  Tutorial, review, and perspectives on open problems,'' 2020.

\bibitem{10.1007/978-3-540-74690-4_12}
D.~Schneega{\ss}, S.~Udluft, and T.~Martinetz, ``Improving optimality of neural
  rewards regression for data-efficient batch near-optimal policy
  identification,'' in \emph{Artificial Neural Networks -- ICANN 2007}, J.~M.
  de~S{\'a}, L.~A. Alexandre, W.~Duch, and D.~Mandic, Eds.\hskip 1em plus 0.5em
  minus 0.4em\relax Berlin, Heidelberg: Springer Berlin Heidelberg, 2007, pp.
  109--118.

\bibitem{4220827}
A.~M. Schaefer, S.~Udluft, and H.-G. Zimmermann, ``A recurrent control neural
  network for data efficient reinforcement learning,'' in \emph{2007 IEEE
  International Symposium on Approximate Dynamic Programming and Reinforcement
  Learning}, 2007, pp. 151--157.

\bibitem{Deisenroth2011c}
M.~P. Deisenroth and C.~E. Rasmussen, ``{PILCO}: A model-based and
  data-efficient approach to policy search,'' in \emph{Proceedings of the
  International Conference on Machine Learning (ICML)}, 2011.

\bibitem{depeweg2017learning}
S.~Depeweg, J.~M. Hern{\'a}ndez-Lobato, F.~Doshi-Velez, and S.~Udluft,
  ``Learning and policy search in stochastic dynamical systems with {Bayesian}
  neural networks,'' in \emph{International Conference on Learning
  Representations}, 2017.

\bibitem{NEURIPS2019_c2073ffa}
A.~Kumar, J.~Fu, M.~Soh, G.~Tucker, and S.~Levine, ``Stabilizing off-policy
  {Q}-learning via bootstrapping error reduction,'' in \emph{Advances in Neural
  Information Processing Systems}, H.~Wallach, H.~Larochelle, A.~Beygelzimer,
  F.~d\textquotesingle Alch\'{e}-Buc, E.~Fox, and R.~Garnett, Eds.,
  vol.~32.\hskip 1em plus 0.5em minus 0.4em\relax Curran Associates, Inc.,
  2019.

\bibitem{wu2019behavior}
Y.~Wu, G.~Tucker, and O.~Nachum, ``Behavior regularized offline reinforcement
  learning,'' 2019.

\bibitem{NEURIPS2020_0d2b2061}
A.~Kumar, A.~Zhou, G.~Tucker, and S.~Levine, ``Conservative {Q}-learning for
  offline reinforcement learning,'' in \emph{Advances in Neural Information
  Processing Systems}, H.~Larochelle, M.~Ranzato, R.~Hadsell, M.~Balcan, and
  H.~Lin, Eds., vol.~33.\hskip 1em plus 0.5em minus 0.4em\relax Curran
  Associates, Inc., 2020, pp. 1179--1191.

\bibitem{NEURIPS2020_a322852c}
T.~Yu, G.~Thomas, L.~Yu, S.~Ermon, J.~Y. Zou, S.~Levine, C.~Finn, and T.~Ma,
  ``{MOPO}: Model-based offline policy optimization,'' in \emph{Advances in
  Neural Information Processing Systems}, H.~Larochelle, M.~Ranzato,
  R.~Hadsell, M.~Balcan, and H.~Lin, Eds., vol.~33.\hskip 1em plus 0.5em minus
  0.4em\relax Curran Associates, Inc., 2020, pp. 14\,129--14\,142.

\bibitem{NEURIPS2020_f7efa4f8}
R.~Kidambi, A.~Rajeswaran, P.~Netrapalli, and T.~Joachims, ``{MOReL}:
  Model-based offline reinforcement learning,'' in \emph{Advances in Neural
  Information Processing Systems}, H.~Larochelle, M.~Ranzato, R.~Hadsell,
  M.~Balcan, and H.~Lin, Eds., vol.~33.\hskip 1em plus 0.5em minus 0.4em\relax
  Curran Associates, Inc., 2020, pp. 21\,810--21\,823.

\bibitem{Swazinna2021}
P.~Swazinna, S.~Udluft, and T.~Runkler, ``Overcoming model bias for robust
  offline deep reinforcement learning,'' \emph{Engineering Applications of
  Artificial Intelligence}, vol. 104, p. 104366, 2021.

\bibitem{NEURIPS2021_f29a1797}
T.~Yu, A.~Kumar, R.~Rafailov, A.~Rajeswaran, S.~Levine, and C.~Finn, ``{COMBO}:
  Conservative offline model-based policy optimization,'' in \emph{Advances in
  Neural Information Processing Systems}, M.~Ranzato, A.~Beygelzimer,
  Y.~Dauphin, P.~Liang, and J.~W. Vaughan, Eds., vol.~34.\hskip 1em plus 0.5em
  minus 0.4em\relax Curran Associates, Inc., 2021, pp. 28\,954--28\,967.

\bibitem{rigter2022ramborl}
M.~Rigter, B.~Lacerda, and N.~Hawes, ``{RAMBO}-{RL}: Robust adversarial
  model-based offline reinforcement learning,'' in \emph{Advances in Neural
  Information Processing Systems}, A.~H. Oh, A.~Agarwal, D.~Belgrave, and
  K.~Cho, Eds., 2022.

\bibitem{SWAZINNA202219}
P.~Swazinna, S.~Udluft, D.~Hein, and T.~Runkler, ``Comparing model-free and
  model-based algorithms for offline reinforcement learning,''
  \emph{IFAC-PapersOnLine}, vol.~55, no.~15, pp. 19--26, 2022, 6th IFAC
  Conference on Intelligent Control and Automation Sciences ICONS 2022.

\bibitem{swazinna2023userinteractive}
P.~Swazinna, S.~Udluft, and T.~Runkler, ``User-interactive offline
  reinforcement learning,'' in \emph{The Eleventh International Conference on
  Learning Representations}, 2023.

\bibitem{abdolmaleki2021multiobjective}
A.~Abdolmaleki, S.~H. Huang, G.~Vezzani, B.~Shahriari, J.~T. Springenberg,
  S.~Mishra, D.~TB, A.~Byravan, K.~Bousmalis, A.~Gyorgy, C.~Szepesvari,
  R.~Hadsell, N.~Heess, and M.~Riedmiller, ``On multi-objective policy
  optimization as a tool for reinforcement learning,'' 2021.

\bibitem{hong2023confidenceconditioned}
J.~Hong, A.~Kumar, and S.~Levine, ``Confidence-conditioned value functions for
  offline reinforcement learning,'' in \emph{The Eleventh International
  Conference on Learning Representations}, 2023.

\bibitem{Deisenroth2011d}
M.~P. Deisenroth and D.~Fox, ``Multiple-target reinforcement learning with a
  single policy,'' ICML 2011 Workshop on Planning and Acting with Uncertain
  Models, 2011.

\bibitem{bischoff2013learning}
B.~Bischoff, D.~Nguyen-Tuong, T.~Koller, H.~Markert, and A.~Knoll, ``Learning
  throttle valve control using policy search,'' in \emph{Machine Learning and
  Knowledge Discovery in Databases: European Conference, ECML PKDD 2013,
  Prague, Czech Republic, September 23-27, 2013, Proceedings, Part I 13}.\hskip
  1em plus 0.5em minus 0.4em\relax Springer, 2013, pp. 49--64.

\bibitem{PIANOSI201110579}
F.~Pianosi, X.~{Quach Thi}, and R.~Soncini-Sessa, ``Artificial neural networks
  and multi objective genetic algorithms for water resources management: an
  application to the {Hoabinh} reservoir in {Vietnam},'' \emph{IFAC Proceedings
  Volumes}, vol.~44, no.~1, pp. 10\,579--10\,584, 2011, 18th IFAC World
  Congress.

\bibitem{5874921}
A.~Castelletti, F.~Pianosi, and M.~Restelli, ``Multi-objective fitted {Q}
  iteration: Pareto frontier approximation in one single run,'' in \emph{2011
  International Conference on Networking, Sensing and Control}, 2011, pp.
  260--265.

\bibitem{10.2166/hydro.2013.169}
F.~Pianosi, A.~Castelletti, and M.~Restelli, ``Tree-based fitted {Q}-iteration
  for multi-objective {Markov} decision processes in water resource
  management,'' \emph{Journal of Hydroinformatics}, vol.~15, no.~2, pp.
  258--270, 01 2013.

\bibitem{pmlr-v37-schaul15}
T.~Schaul, D.~Horgan, K.~Gregor, and D.~Silver, ``Universal value function
  approximators,'' in \emph{Proceedings of the 32nd International Conference on
  Machine Learning}, ser. Proceedings of Machine Learning Research, F.~Bach and
  D.~Blei, Eds., vol.~37.\hskip 1em plus 0.5em minus 0.4em\relax Lille, France:
  PMLR, 07--09 Jul 2015, pp. 1312--1320.

\bibitem{PARISI20173}
S.~Parisi, M.~Pirotta, and J.~Peters, ``Manifold-based multi-objective policy
  search with sample reuse,'' \emph{Neurocomputing}, vol. 263, pp. 3--14, 2017,
  multiobjective Reinforcement Learning: Theory and Applications.

\bibitem{pong2018temporalICLR}
V.~Pong, S.~Gu, M.~Dalal, and S.~Levine, ``Temporal difference models:
  Model-free deep {RL} for model-based control,'' in \emph{International
  Conference on Learning Representations}, 2018.

\bibitem{plappert2018multigoal}
M.~Plappert, M.~Andrychowicz, A.~Ray, B.~McGrew, B.~Baker, G.~Powell,
  J.~Schneider, J.~Tobin, M.~Chociej, P.~Welinder, V.~Kumar, and W.~Zaremba,
  ``Multi-goal reinforcement learning: Challenging robotics environments and
  request for research,'' 2018.

\bibitem{ijcai2022p0770}
M.~Liu, M.~Zhu, and W.~Zhang, ``Goal-conditioned reinforcement learning:
  Problems and solutions,'' in \emph{Proceedings of the Thirty-First
  International Joint Conference on Artificial Intelligence, {IJCAI-22}}, L.~D.
  Raedt, Ed.\hskip 1em plus 0.5em minus 0.4em\relax International Joint
  Conferences on Artificial Intelligence Organization, 7 2022, pp. 5502--5511,
  survey Track.

\bibitem{RL_Unplugged}
C.~Gulcehre, Z.~Wang, A.~Novikov, T.~Paine, S.~G\'{o}mez, K.~Zolna, R.~Agarwal,
  J.~S. Merel, D.~J. Mankowitz, C.~Paduraru, G.~Dulac-Arnold, J.~Li,
  M.~Norouzi, M.~Hoffman, N.~Heess, and N.~de~Freitas, ``{RL} unplugged: A
  suite of benchmarks for offline reinforcement learning,'' in \emph{Advances
  in Neural Information Processing Systems}, H.~Larochelle, M.~Ranzato,
  R.~Hadsell, M.~Balcan, and H.~Lin, Eds., vol.~33.\hskip 1em plus 0.5em minus
  0.4em\relax Curran Associates, Inc., 2020, pp. 7248--7259.

\bibitem{D4RL}
\BIBentryALTinterwordspacing
J.~Fu, A.~Kumar, O.~Nachum, G.~Tucker, and S.~Levine, ``{D4RL}: Datasets for
  deep data-driven reinforcement learning,'' 2021. [Online]. Available:
  \url{https://openreview.net/forum?id=px0-N3_KjA}
\BIBentrySTDinterwordspacing

\bibitem{NEORL}
R.-J. Qin, X.~Zhang, S.~Gao, X.-H. Chen, Z.~Li, W.~Zhang, and Y.~Yu, ``{NeoRL}:
  A near real-world benchmark for offline reinforcement learning,'' in
  \emph{Advances in Neural Information Processing Systems}, S.~Koyejo,
  S.~Mohamed, A.~Agarwal, D.~Belgrave, K.~Cho, and A.~Oh, Eds., vol.~35.\hskip
  1em plus 0.5em minus 0.4em\relax Curran Associates, Inc., 2022, pp.
  24\,753--24\,765.

\bibitem{openai_gym}
G.~Brockman, V.~Cheung, L.~Pettersson, J.~Schneider, J.~Schulman, J.~Tang, and
  W.~Zaremba, ``Openai gym,'' 2016.

\bibitem{hein2017batch}
D.~Hein, S.~Udluft, M.~Tokic, A.~Hentschel, T.~A. Runkler, and V.~Sterzing,
  ``Batch reinforcement learning on the industrial benchmark: First
  experiences,'' in \emph{2017 International Joint Conference on Neural
  Networks (IJCNN)}.\hskip 1em plus 0.5em minus 0.4em\relax IEEE, 2017, pp.
  4214--4221.

\end{thebibliography}

\end{document}